\documentclass[11pt,onecolumn]{IEEEtran}
\usepackage{cite}
\usepackage[left=4cm, right=4cm, top=4cm]{geometry}
 
\ifCLASSINFOpdf
\else
\fi
\ifCLASSOPTIONcompsoc
 \usepackage[font=normalsize,labelfont=sf,textfont=sf]{subfig}
\else
 \usepackage[font=footnotesize]{subfig}
\fi
\ifCLASSOPTIONcompsoc
 \usepackage[font=normalsize,labelfont=sf,textfont=sf]{subfig}
\else
 \usepackage[font=footnotesize]{subfig}
\fi

\usepackage{setspace}

\usepackage{color}
 \usepackage{amssymb}
\usepackage{amsfonts}
\usepackage{dsfont}
\usepackage{epsfig}

\usepackage{times}
\usepackage{graphicx}
\usepackage{amsmath}
\usepackage[ruled,lined,linesnumbered,boxed]{algorithm2e}

\usepackage{cleveref}

\def\G{\mathcal{G}}

\def\V{\mathcal{V}}

\DeclareMathSymbol{\R}{\mathbin}{AMSb}{"52}

\newcommand\bX{\bold{X}}
\newcommand\bY{\bold{Y}}
\newcommand\bK{\bold{k}}
\newcommand\bC{\bold{C}}
\newcommand\bPi{\bold{\pi}}

\def\by#1{\mathop{{\hbox{\setbox0=\hbox{$\scriptstyle{#1\quad}$}{$
\mathrel{\mathop{\setbox1=\hbox to \wd0{\rightarrowfill}\ht1=3pt\dp1=-2pt\box1}\limits^{#1}}
$}}}}}

\title{Finite State Machines for Semantic Scene Parsing and Segmentation}

\author{Hichem Sahbi \\ $ $ \\ {CNRS, LIP6 Lab, Sorbonne University, Paris}}

\begin{document}

\maketitle

\begin{abstract}
We introduce in this work a novel stochastic inference process, for scene annotation and object class segmentation, based on finite state  machines (FSMs). The design principle of our framework is generative and based on building, for a given scene, finite state machines that encode annotation lattices, and inference consists in finding and scoring the best configurations in these lattices. Different novel operations are defined using our FSM framework including reordering, segmentation, visual transduction, and label dependency modeling. All these operations are combined together in order to achieve annotation as well as object class segmentation.  
\end{abstract}


\section{Introduction}
\label{sec:intro}
The general problem of image annotation is usually converted into  classification. Many existing state of the art methods~(see for instance \cite{Carneiro-Vasconcelos-CVPR05,sahbi2003coarse,tollari2008comparative,He2004,Sin2003,bourdis2011constrained,Nowak2010,boujemaa2004visual,ref7,napoleon20102d,Jeon-Lavrenko-Manmatha-sigir2003,ref18,sahbi2002coarse,ref14,ref8,Liu-Li-Liu-pr2009,Sahbi1}) treat each keyword (also referred to as label or category) as a class, and then build the corresponding category-specific classifier in order to identify images or image regions  belonging to that class, using a variety of machine learning techniques such as Markov models~\cite{Li-Wang-PAMI03,Moser2012,li2000image,choi2001multiscale}, latent Dirichlet allocation~\cite{Barnard-Duygululu-Forsyth-JMLR03,blei2003latent,rasiwasia2013latent,wang2008spatial}, probabilistic latent semantic analysis~\cite{Monay-GaticaPerez-acmmm04,jin2012image,zhong2015scene}, support vector machines~\cite{Sahbi6,kim2002support,Gao-Fan-Xue-Jain-acmmm06,sahbi2015imageclef,wang2011color,vo2012transductive,Moser2012,sahbi2013cnrs,Semo2010}, deep learning~\cite{lecun2015deep,jiu2017nonlinear,ChenPKMY18,krizhevsky2012imagenet,szegedy2015going,sahbi2017coarse,he2016deep,resnet2015,badrinarayanan2017segnet,jiu2016deep}, etc. Annotation methods may also be categorized into: {\it  region-based} object class segmentation (OCS)  requiring a preliminary step of image segmentation~(e.g., \cite{Ladick-Russell-Kohli-ICCV2009,Pantofaru-Schmid-Hebert-ECCV2008,Reynolds-Murphy-CRV2007,Sahbi10,Girshick15,Shotton-Johnson-Cipolla-CVPR2008,Batra-Sukthankar-Tsuhan-CVPR2008,Galleguillos-Rabinovich-Belongie-CVPR2008,Yang-Meer-Foran-CVPR2007,FelzenszwalbH04,Shotton-Winn-Rother-A-Criminisi-ECCV2006,Kohli-Ladick-Torr-CVPR2008}, etc.), and {\it holistic}~(e.g., \cite{sahbi2008robust,Jeon-Lavrenko-Manmatha-sigir2003,jiu2016laplacian,Wang-Yan-Zhang-Zhang-cvpr2009,Sahbi1,fcn2015,unets15,Sahbi6,long2015fully}, etc.) operating directly on the entire image space. In both cases, training is achieved in order to learn how to attach labels to the corresponding visual features. \\

\indent In this paper, we introduce an original OCS method based on FSMs. Our approach is Bayesian; it finds  superpixel labels that maximize a posterior probability, {\it but} its key-novelty resides in the representational power of FSMs in order to build a  comprehensive model for scene segmentation and labeling.  Indeed, we translate our  OCS into searching, via FSMs, the optimum of a discrete energy function mixing i) a  {\it unary term} that models conditional probability of visual features given their (possible) labels, ii) an {\it  interaction potential}  which provides  joint statistics, of co-occurrence, between those labels, and more importantly iii) a novel {\it reordering and grouping term}. The latter allows us, via FSMs, to examine (generate, label, score and rank) many partitions of segments of a scene, and to return only the likely ones. \\

\noindent  At least two reasons drove us to apply FSMs for OCS:\\

\noindent i) Firstly, as FSMs can model huge (even infinite) languages\footnote{\scriptsize In OCS, the alphabet of the language corresponds to all the superpixels of a given scene.}, with finite    memory and time resources, our  method does not require explicit generation of all  possible  segmentations and labelings. Instead, it first models them implicitly by combining (composing) different FSMs and then efficiently finds the shortest path (i.e., the most likely segmentation and labeling) in a global FSM.\\

\noindent ii) Secondly, superpixel reordering allows us to examine the possible segmentations at different orders and, using the chain rule, to maximize the interaction potentials resulting into better labeling. For that purpose,  scenes are first described with  graphs where nodes correspond to superpixels and edges connect neighboring superpixels. Then, reordering is achieved by  generating  random (Hamiltonian) walks on these graphs using FSMs. Note that  graphs with very low connectivity ($\leq 4$ immediate neighbors for each superpixel) dramatically reduce  the complexity of this reordering and also grouping while, at the same time, make it possible to explore larger sets of possible solutions resulting into an effective and also efficient scene labeling machinery as discussed later in this paper.
\section{Scene Labeling Model} 
Given $n$ lattice points ${\cal V}=\{1,\dots,n\}$, we define in this section $\bX=\{X_1,\dots,X_n\}$ as a set of {\it observed} random variables, corresponding to a subdivision of an image $\bX$ into smaller units,  referred to as {\it superpixels}.  Let $\bC=\{c_i: c_i \subseteq \V\}_{i=1}^\bK$ be  a {\it random partition} of $\V$; an element $\bX_{c_i} \subseteq \bX$ is defined as a collection of conditionally independent random variables (here $\bX_{c_i}=\{X_k \in \bX: k \in c_i\}$) and $\bY=\{\bY_{c_1},\dots,\bY_{c_\bK}\}$ the underlying (unknown) labels taken from a label set ${\cal C} = \{\ell_i\}_i$. \\ 
For a given observed superpixel set $\bX$, our scene labeling defines a joint probability distribution over multiple superpixel {\it reorderings}, {\it groupings} (segmentations), {\it labelings} and finds the best tuple $(\bY,\bC,\bK,\bPi)$ as the max of the following probability distribution 
\begin{eqnarray}\label{eq1}
\scriptsize 
 P(\bX, \bY,\bC, \bK,{\bf \bPi})   = &  \\ 
P(\bPi).  & \textrm{\small Superpixel Reordering} \\ 
P(\bC,\bK | \bPi).  & \textrm{\small Superpixel Grouping Model} \\ 
  P(\bY | \bC,\bK,\bPi).   &    \textrm{\small Label Dependency Model} \\
  P(\bX |\bY,\bC,\bK, \bPi).   & \textrm{\small Visual Model.}
\end{eqnarray} 

\begin{figure*}[hptb]
\begin{center}
\hspace{-1.75cm}\scalebox{0.8}{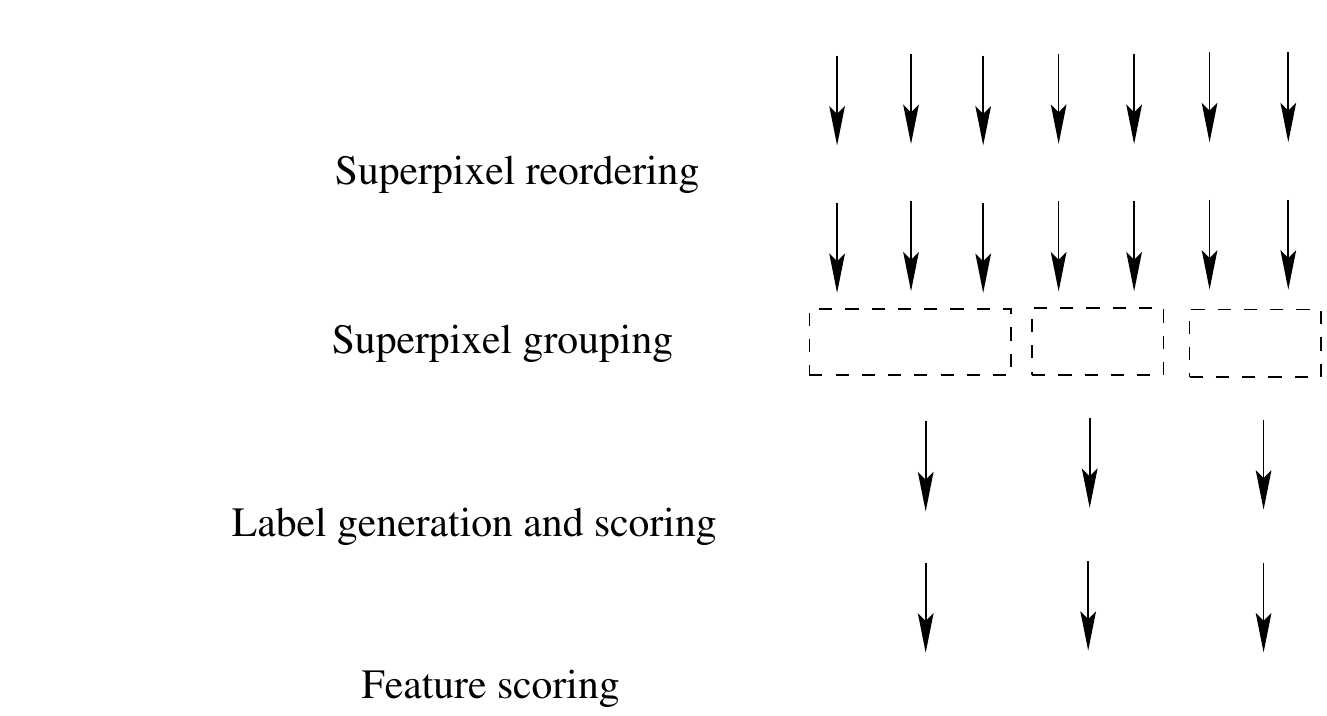}   
\end{center}
\vspace{.5cm}
\begin{center}
\scalebox{0.7}{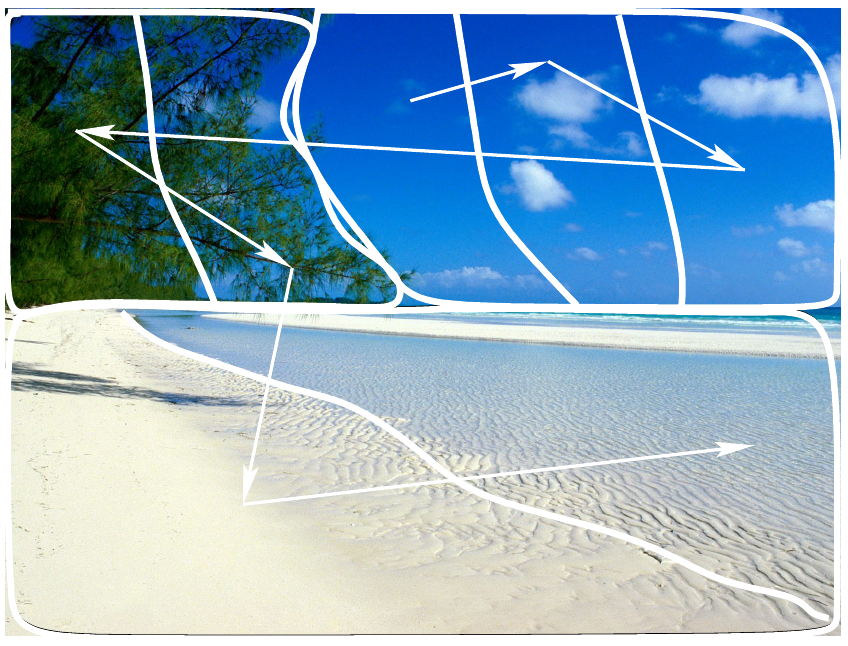}\hspace{0.15cm}\scalebox{0.7}{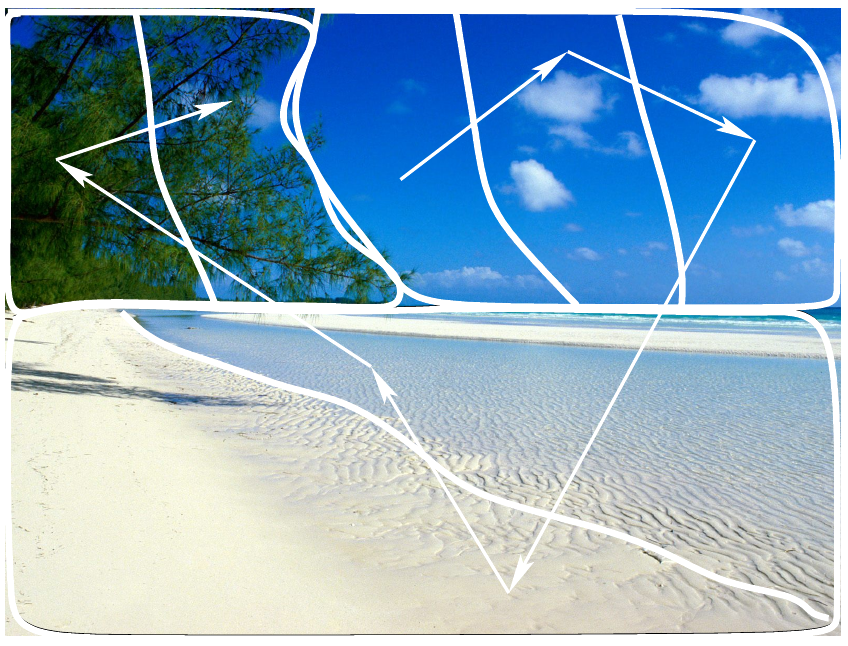}
\end{center}
\caption{(Top) This figure shows one realization of the stochastic image labeling process. (Bottom) This figure shows two parsing possibilities corresponding to two different superpixel permutations; this example illustrates the principle from reordering three segments $c_1=\{s_1,s_2,s_3\}$, $c_2=\{s_4,s_5\}$ and $c_3=\{s_6,s_7\}$. If the underlying labels ($\bY_{c_1}=\text{sky}$, $\bY_{c_2}=\text{tree}$)  are less likely to co-occur than ($\bY_{c_1}=\text{sky}$, $\bY_{c_3}=\text{sea}$), then one should reorder them as $c_1, c_3,c_2$ prior to estimate their dependency statistics using the chain rule; i.e., assuming a $1^{\textrm{st}}$ order Markov process, one should consider  $P(\bY_{c_1}). P(\bY_{c_3}|\bY_{c_1}).P(\bY_{c_2}|\bY_{c_3})$ instead of  $P(\bY_{c_1}). P(\bY_{c_2}|\bY_{c_1}).P(\bY_{c_3}|\bY_{c_2})$.   Note also that parsing images should not be achieved as 1D patterns (such as speech or text) since the order in images is obviously not unique.}\label{fig:realization}
\end{figure*} 

\noindent Here $\bPi \in {\cal G}(\V)$ denotes a permutation (reordering) that maps each element $i\in\V$ to $\bPi_i \in \V$ and  ${\cal G}(\V)$ denotes the symmetric group on $\V$ including all the bijections (permutations) from $\V$ to it-self. The model above illustrates a generative scene labeling process which first (i) reorders ({\it in multiple ways}) the lattice $\V$ resulting into $\bPi_1\dots\bPi_n$, (ii) partitions/groups ({\it in multiple ways}) $\bPi_1\dots\bPi_n$ into $\bK$ subsets $c_1\dots c_\bK$, then (iii) emits ({\it in multiple ways}) label hypotheses  $\bY_{c_1}\dots\bY_{c_\bK}$ for the segments $\bX_{c_1}\dots\bX_{c_\bK}$ and finally (iv) estimates their visual likelihood so only relevant labels will be strengthened. Example in Fig.~(\ref{fig:realization}, top) illustrates one realization of this stochastic process.\\
Note that a naive and brute force generation of all the possible reorderings, groupings, labelings would be out of hand; as detailed in the subsequent sections,  our approach does not explicitly generate  all these configurations,  but instead {\it implicitly specifies them using finite state machines} in order to build a scored lattice of possible segmentations and labelings.

\subsection{Reordering \& Grouping Models for Multiple Image Segmentation}\label{reorderingandgrouping}

Reordering corresponds to  permutations that transform an image lattice $\V$ into many words  in $\{\bPi\}_{\bPi \in {\cal G}(\V)}$, while grouping breaks every word $\bPi=\bPi_1\dots\bPi_n$ into $\bf k$ subwords. Applying all the permutations $\{\bPi\}_{\bPi  \in \G(\V)}$, followed by all the possible groupings makes it possible to generate all the possible partitions of $\V$. Subsets in each partition (again denoted $\bC=\{c_i\}_{i=1}^{\bf k}$) are defined as $c_i=\{\pi_{k_i},\dots,\pi_{\ell_i}\}$, with $1 \leq k_i \leq \ell_i \leq n$. Reordering and grouping models, also shown in Eqs.~2-3, are necessary not only to delimit segment boundaries with a high precision but also to evaluate label dependencies  between segments  at multiple orders (see section~\ref{sec:dependencymodel} and Fig.~\ref{fig:realization}, bottom). \\
\noindent In this work, we restrict image partitions to include only connected segments with homogeneous superpixels. For that purpose  our superpixel grouping follows random walks (RWs): it randomly visits superpixels in $G=(V,E)$ (graph associated to the lattice $\cal V$), and groups (with a high probability) {\it only} connected and visually similar ones. Each RW corresponds to a path in $G$ which is not necessarily Hamiltonian.  If one restricts RWs to include only permutations, then the resulting paths will be Hamiltonian\footnote{\scriptsize Note that planar 4-connected graphs (including 4-connected regular grids) are necessarily Hamiltonian.} and correspond to partitions of $\V$ that necessarily include connected subsets. \\ 

Considering the lattice $\V$, our {\it reordering model}  $P(\bPi)$ equally weights permutations in $\G(\V)$, i.e., $\forall \bPi \in \G(\V)$,  $P(\bPi) =1\slash{{|{\cal G}(\V)|}}$ while our {\it grouping model} $P(\bC,\bK| \bPi)$  assumes all segments in a given partition conditionally independent given $\bPi$, so one may write $P(\bC,\bK| \bPi) = P(\bK) \prod_{i=1}^{\bK} P(c_i|\bPi)$.  All the partition sizes have the same mass, i.e., $P(\bK) = 1/n$, $\forall  \ \bK \in \{1,2,\dots,n\}$ and  $P(c_{i}|\bPi)  =  P(\bPi_{k_i}) \prod_{j=1}^{\ell_i-k_i} P(\bPi_{k_i+j} | \bPi_{k_i+j-1})$. Here  $P(\bPi_{k_i})$ is taken as uniform, i.e., $1\slash n$ and    $P(\bPi_{k_i+j}|\bPi_{k_i+j-1})$ is taken as the random walk transition probability from  node (superpixel) $\bPi_{k_i+j-1}$ to node $\bPi_{k_i+j}$, which is positive only if the underlying superpixels share a common boundary and it is set to $P(\bPi_{k_i+j}|\bPi_{k_i+j-1})  \ \propto \  \mathds{1}_{\{(\bPi_{k_i+j},\bPi_{k_i+j-1}) \in E \}} \cdot {\bf \kappa}\big(\psi(\bPi_{k_i+j}),\psi(\bPi_{k_i+j-1})\big)$; here ${\bf  \kappa}$ is the histogram intersection kernel and $\psi(\bPi_{k_i+j})$ denotes a visual feature extracted at superpixel  $\bPi_{k_i+j}$. With this model, if the transition  between neighboring superpixels is achieved with a high conditional probability, then these superpixels are considered as visually similar and {\it likely} to come from the same physical object. 

\subsection{Visual and Label Dependency Models}\label{sec:dependencymodel}
Once superpixels reordered and grouped in multiple ways, we use a unary visual and label interaction models, described below,  in order to score the resulting partitions. As shown in the remainder of this paper, only highly scored partitions are likely to correspond to correct object segmentations. \\

\noindent {\bf Label Dependency Model.} This model captures scene structure and a priori knowledge about segment/label relationships (either co-occurrence or geometric relationships) in order to consolidate labels which are consistent with already observed scenes. Considering a first order Markov process and using the chain rule, our bi-gram label dependency model is   $P(\bY | \bC,\bK,\pi) =  P(\bY_{c_1})\prod_{i=2}^{\bK}   P(\bY_{c_i}|\bY_{c_{i-1}})$. \\ 

 Let ${\cal I}=\{{\bf I}_1,\dots,{\bf I}_N\}$ be a training set, of fixed size images, labeled at the pixel level (with labels in $\cal C$). Let ${\bf f}_u(\ell,p)=\sum_{i=1}^N \mathds{1}_{\{{\bf I}_{i}(p)=\ell\}}$ be the frequency of co-occurrence of pixel $p$ and label $\ell$ in  $\cal I$. 
Similarly, we define ${\bf f}_b(\ell,\ell',p,p')$ as  $\sum_{i=1}^N \mathds{1}_{\{{\bf I}_{i}(p)=\ell\}} \mathds{1}_{\{{\bf I}_{i}(p')=\ell'\}}$. Given superpixels $s$, $s'$ with labels resp. as ${\bY}_s$, ${\bf Y}_{s'}$, we define
 \begin{equation}
\begin{array}{lll}
\displaystyle P(\bY_{{s}}) & = &  \displaystyle \frac{\sum_{p \in s}{\bf f}_u(\bY_s,p)}{\sum_{\ell \in {\cal C}} \sum_{p\in s} {\bf f}_u(\ell,p)} \\
& &  \\
 \displaystyle P(\bY_s | \bY_{s'}) & = & \displaystyle  \frac{\sum_{p\in s}\sum_{p'\in s'}{\bf f}_b({\bf Y}_s, \bY_{s'},p,p')}{\sum_{\ell \in {\cal C}} \sum_{p\in s}\sum_{p'\in  s'} {\bf f}_b(\ell, \bY_{s'},p,p')}.
\end{array}
\end{equation} 

\noindent {\bf Visual Model.} This defines the likelihood of  ${\bf X}=\{{\bf X}_{c_1},\dots,{\bf X}_{c_{\bf  k}}\}$ given the labels ${\bf Y}=\{{\bf Y}_{c_1},\dots,{\bf Y}_{c_{\bf  k}}\}$. Assuming conditionally independent superpixels and segments given their labels, and assuming that each ${\bf X}_{c_i}$ depends only on ${\bf Y}_{c_i}$, we define our visual model as 
\begin{equation}
\begin{array}{lll} 
\displaystyle  P({\bf X} | {\bf Y},{\bf C},{\bf k}, {\bf \pi}) & = &  \displaystyle  \displaystyle \prod_{i=1}^\bK  \prod_{s\in c_i} P({\bf X}_{s} | \bY_{c_i}) \\
 & & \\
\displaystyle \textrm{with} \ \ \  \ P({\bf X}_s|{\bf Y}_{c_i})  &\propto&  \displaystyle \frac{1}{1+\exp(-f_{\bY_{c_i}}({\bf X}_s))},
\end{array}
\end{equation}  
\noindent here  $f_{\bY_{c_i}}(.)$ is an SVM classifier (based on histogram intersection kernel) trained, using LIBSVM, to discriminate  superpixels belonging to a category $\bY_{c_i}$ from ${\cal C}\backslash \bY_{c_i}$.

\section{Finite State Machine Inference}\label{sec:fsm} 

In this section, we implement the models discussed earlier using FSMs. We  remind, in~\cite{sahbi2014NIPS}, the definition of stochastic FSMs\footnote{\scriptsize All the definitions and implementations of FSMs are reported in the technical report~\cite{sahbi2014NIPS}.}, in particular finite state acceptors (FSA) and transducers (FST), and we show how we design and combine those machines in order to build a global transducer. The latter encodes in a compact way, the lattice of possible annotations of a given scene and the best annotation corresponds to the best path in that lattice. \\ 

\begin{figure}[thp]
\begin{center}
\hspace{-0.8599999cm}\includegraphics[angle=-90,width=1.0\linewidth]{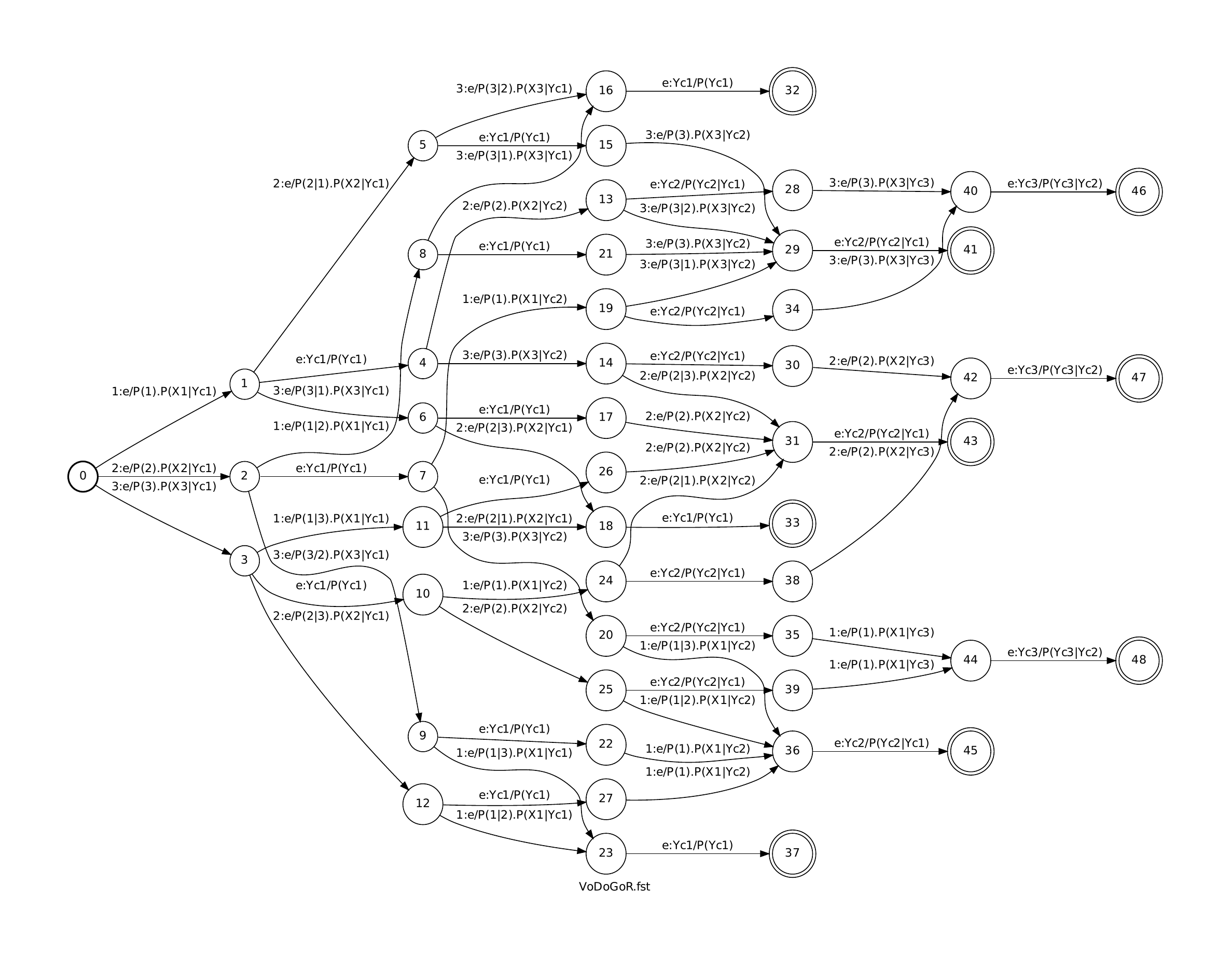}
\end{center}
\caption{This figure shows an example of the global FSM ${\bf V} \circ {\bf D} \circ {\bf G} \circ {\bf R}$. {(\bf Better to view the zoomed version of the PDF)}}
\label{fig:superpixel_classification2}
\end{figure}

Given a scene paved with a collection of non-overlapping superpixels, our labeling model first  reorders these superpixels (via a ``reordering  FSA'' $\bf R$)  and  group them (via a ``grouping FST''  $\bf G$). These two FSMs when composed together, allow us to generate many reordered partitions of segments\footnote{\scriptsize Each segment in these partitions is seen as a ``phrase'' in a ``language'' with an alphabet corresponding to  all the superpixels of a scene.}. Among these  partitions, only a few of them are relevant and  correspond to {\it meaningful} objects in the scene. Therefore, and in order to find  these relevant partitions, we combine the $\bf R$ and $\bf G$ FSMs with ${\bf V} \circ {\bf D}$ (resulting from the composition of visual and label dependency FSTs; see example in~\cite{sahbi2014NIPS}) that scores segments in all possible partitions, depending on their {\it unary} and {\it binary} interactions, and returns only the most likely partition and its labels. The most likely solution (partition and its labels) corresponds to the best (shortest) path in the global FSM (${\bf V} \circ {\bf D} \circ {\bf G} \circ {\bf R}$)  provided that negative log-likelihood transform is applied to all FSM transition probabilities; this solution also minimizes the energy  $- \log P({\bf X}, {\bf Y}, {\bf C}, {\bf k},\pi)$ (see Eq.~1).  Figure~\ref{fig:superpixel_classification2} shows an example of this global FSM. \\ 

\indent Note that the global composition process (${\bf F}={\bf V} \circ {\bf D} \circ  {\bf G} \circ {\bf R}$) could be time and memory demanding. One may significantly reduce complexity of  different transducers (and thereby the global one) at different levels including the visual and the label dependency models by only keeping sparse transitions (related to strictly positive or large statistics). Another simplification consists in reducing the number of possible labels, especially if one is interested in domain specific applications with restricted labels. In practice, with these simplifications (and besides superpixel and feature extraction), FSM inference takes $<$ 2s to annotate an image of $20 \times 20$ superpixels using a standard 3Ghz PC.

\begin{table*}
\centering
  \resizebox{0.99\linewidth}{!}{
\begin{tabular}{cccc||ccc} 
  \ \   Lines &  Reordering + Grouping Model              & Label Dependency Model   &  Visual Model      & FAR (\%) & FRR (\%) & EER (\%)  \\ 
\hline  
\hline 
1: & Yes/No                                 &  flat              &  flat         &   25.00 & 75.00 & 50.00 \\ 
\hline
2: & No                                &  flat              &  learned       & 04.98   & 28.81  &  16.90  \\
3: & No                                &  learned            &  flat         & 13.47    &  41.26  & 27.36  \\ 
4: & No                                &  learned            &  learned       & {\bf 03.11}   &  {\bf 11.34}  & {\bf 07.22}   \\
\hline
5: & Yes + flat                         &  flat              &  learned       &   04.98  & 28.78 & 16.88 \\ 
6: & Yes + flat                          &  learned            &  flat         &   11.43 & 36.52 & 23.98  \\ 
7: & Yes + flat                         &  learned            &  learned       &   {\bf 02.70}  & {\bf 08.64}  & {\bf 05.67}  \\ 
\hline
8: & Yes + learned                      &  flat              &  learned       &   04.98  & 28.96  & 16.97 \\
9: & Yes + learned                       &  learned            &  flat         &   11.13 & 33.99 & 22.56  \\
10: & Yes + learned                       &  learned            &  learned       &   {\bf 02.49}  & {\bf 07.80} & {\bf 05.14} 
\end{tabular}
}
\caption{This table shows the performance of our model for different settings on the Sunset04 dataset.}\label{tab1} 
\end{table*} 

\section{Validation and Discussion}\label{validation} 

\noindent {\bf Evaluation Set and Setting.} In order to validate the proposed OCS approach, we use the  ``Sunset04'' database. The latter contains 100 sunset scenes with objects belonging to 4 categories  (``Sky'', ``Sea'', ``Sand'' and ``Sun''); note that this database is challenging and interesting for our model  as it has visually confusing categories (such as ``Sky vs. Sea'', etc.). Given an image, the goal is to assign each group of pixels (i.e., superpixel) to one of these 4  categories. In practice, a given image is subdivided into a regular grid of 400 ($20 \times 20$) superpixels, each one is connected to its 4 immediate neighbors: top, bottom, left, right (see other e.g.~\cite{thiemert2005applying,thiemert2006using}) and described using the bag-of-word SIFT representation. Precisely, dense SIFT features are extracted and quantized using a codebook of 200 visual words and a two level spatial pyramid is used to describe each superpixel resulting into a feature vector of 1000 dimensions. \\
For each image in Sunset04, we turn OCS into an FSM decoding process (as described earlier); half of the Sunset04 database is used in order to train (obtain statistics of)  different FSMs (i.e., label dependency and visual models) while the other half is used as a test set for OCS decoding. For each category, we measure the underlying false acceptance rates (FAR) and false rejection rates (FRR) as well as the equal error rates (as average of FAR and FRR) all at the pixel level; we report the mean accuracy as the expectation of these errors through different categories.\\ 

\noindent {\bf Model Evaluation.} Table~\ref{tab1} reports the accuracy of our FSM decoding for different setting of our model. The main goal is to understand the contribution of each step in the FSM decoding/inference process.  In this table, ``Yes'' stands for the use  of the grouping+reordering models; the transition probabilities of the random walk grouping model are either uniform ({\it flat}) or set ({\it learned}) as described in Section II. A ``No'' stands for no-reordering which means that a given test image is parsed using a unique order (lexicographic in practice).  We also consider, in these experiments, {\it flat} and {\it learned} label dependency  and visual models. Flat models mean that all the underlying statistics are uniform while learned models correspond to  statistics taken from Eqs.~6-7. In contrast to flat reordering+grouping models, setting flat label dependency (or visual) model is strictly equivalent to the ``non-use'' of that model as its impact on the decoding process becomes completely neutral.\\ 

\noindent {\bf Impact of Reordering and Grouping Model.} Lines 2-4 vs. 5-10, show that the OCS performances are consistently  better when using reordering (R) and grouping (G) models as the latter parse and group superpixels with different orders. This results into a better exploration of the space of possible solutions of Eq.~1 (segmentations, labelings and dependencies)  and thereby a better OCS performance. Note that learned R and G models (Lines 8-10) always outperform flat ones (Lines 5-7), as the (random walk-based) grouping model gives more preference and better scoring to visually similar superpixels  which are more likely to correspond to actual objects.\\

\begin{figure}
\centering
{\includegraphics[width=0.12\linewidth]{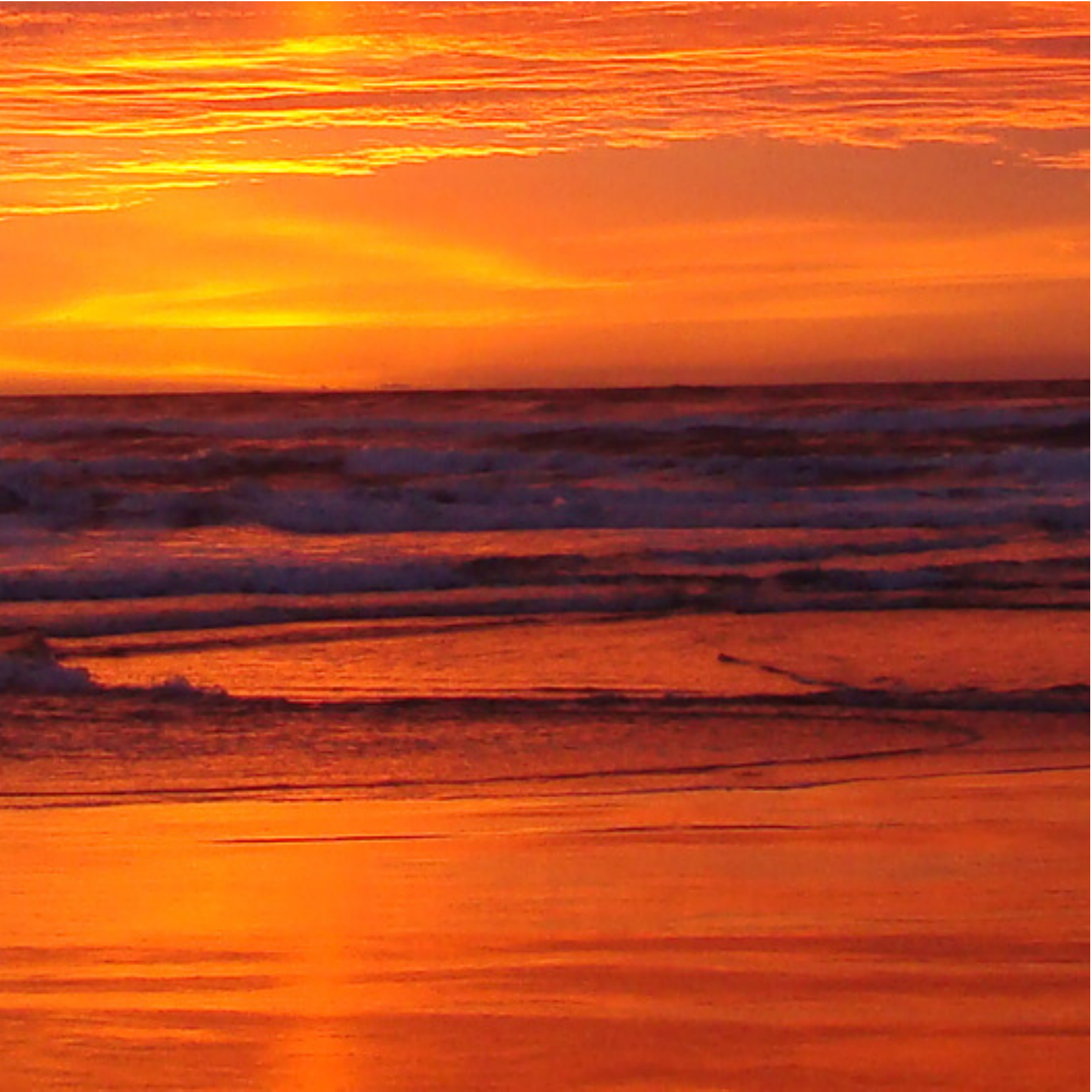}}\,\,
{\includegraphics[width=0.12\linewidth]{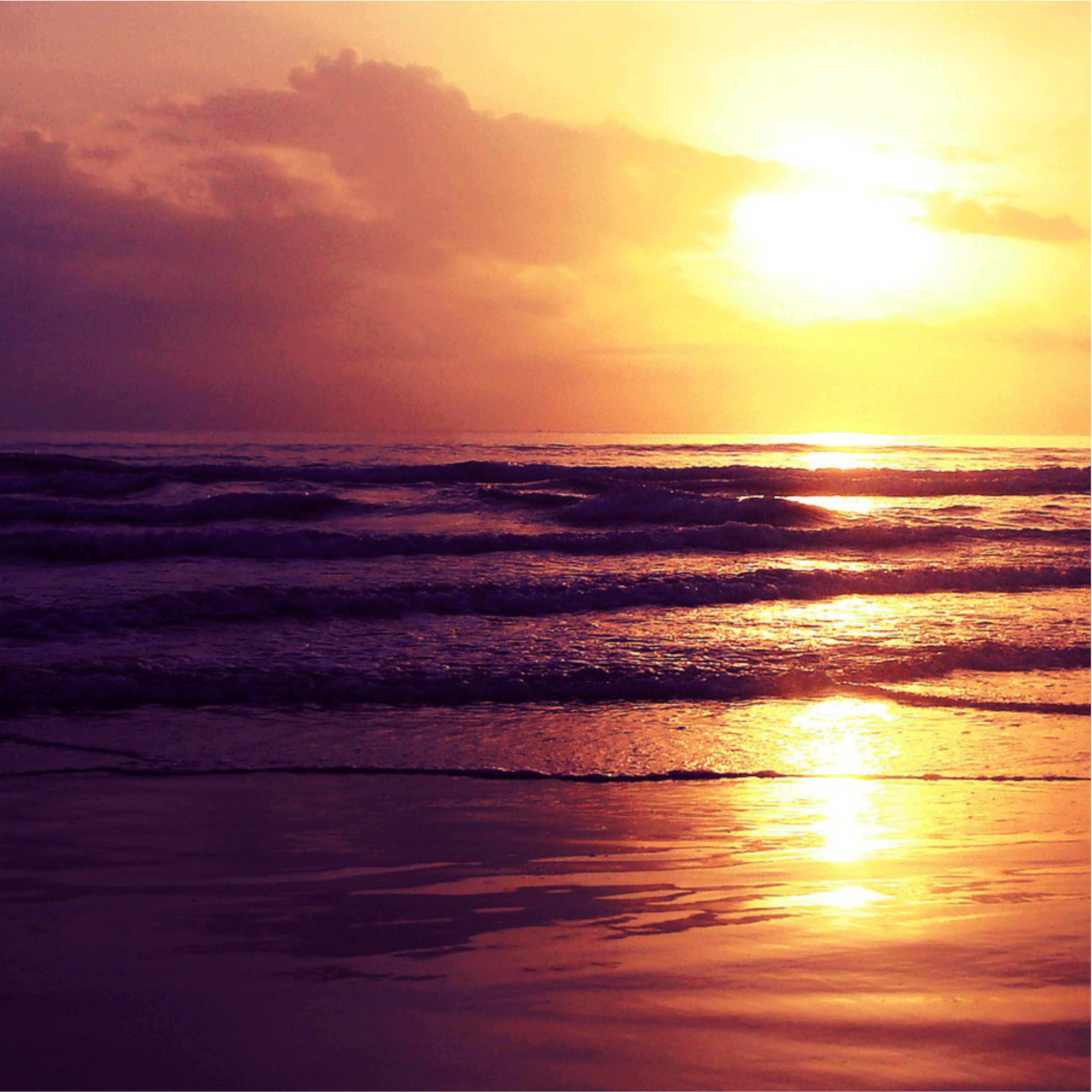}}\,\,
{\includegraphics[width=0.12\linewidth]{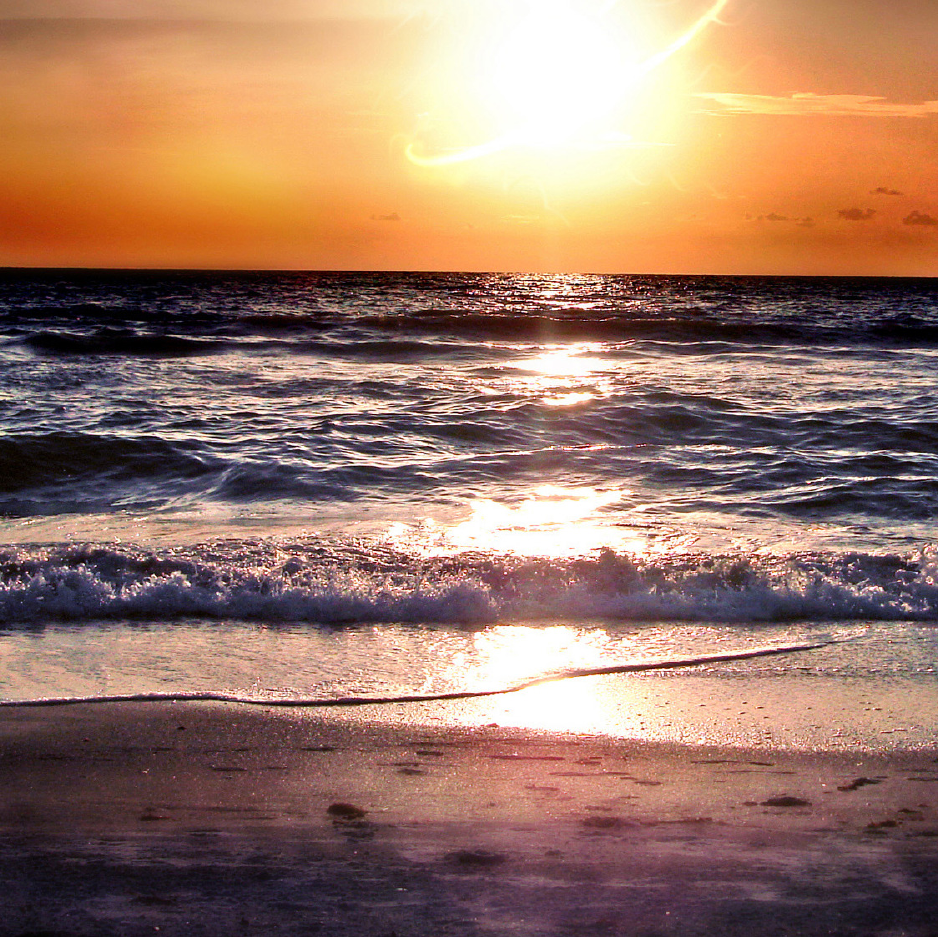}}\,\,
{\includegraphics[width=0.12\linewidth]{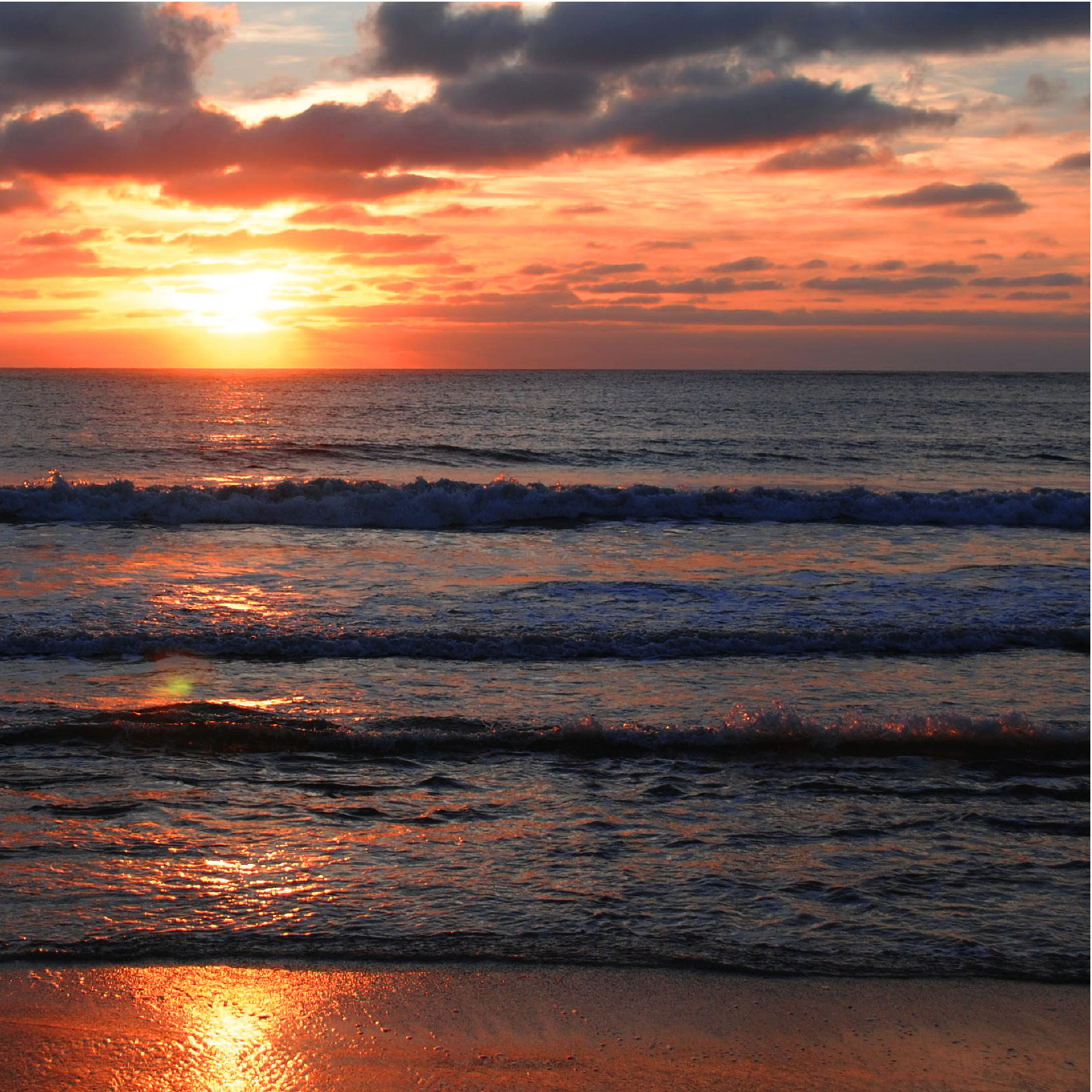}}\,\, \\ \vspace{0.08cm}
{\includegraphics[width=0.12\linewidth]{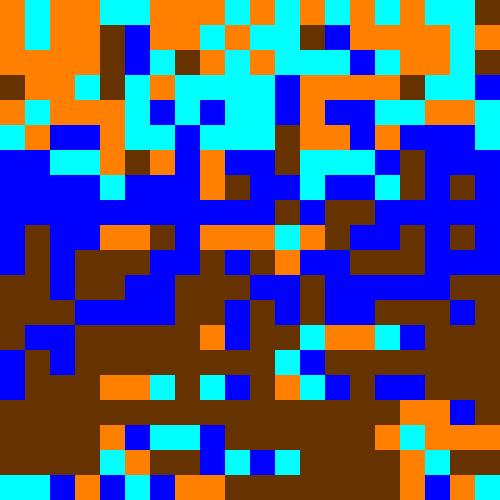}}\,\,
{\includegraphics[width=0.12\linewidth]{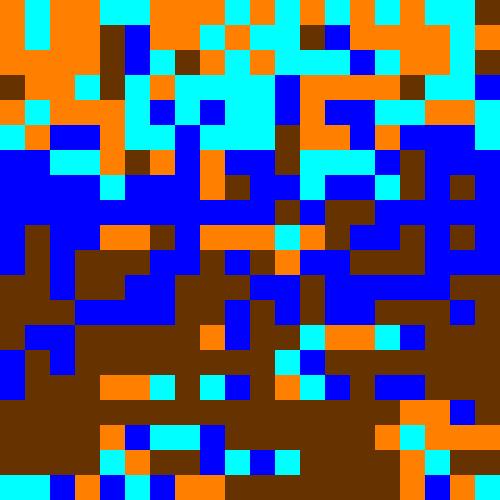}}\,\,
{\includegraphics[width=0.12\linewidth]{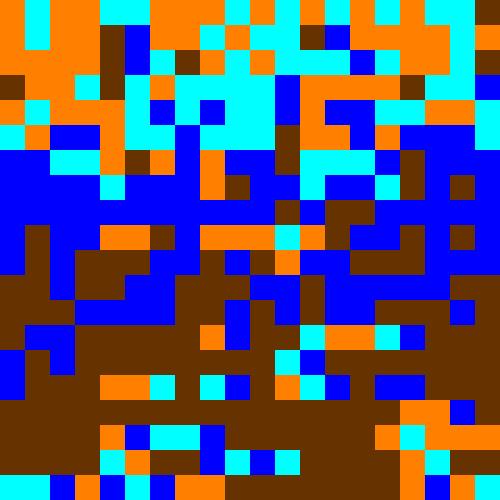}}\,\, 
{\includegraphics[width=0.12\linewidth]{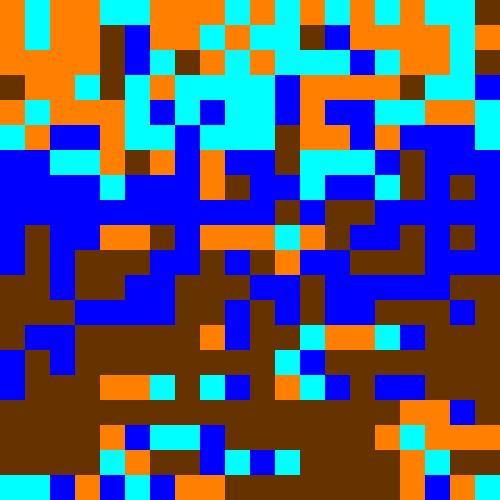}}\,\ \\ \vspace{0.15cm}
{\includegraphics[width=0.12\linewidth]{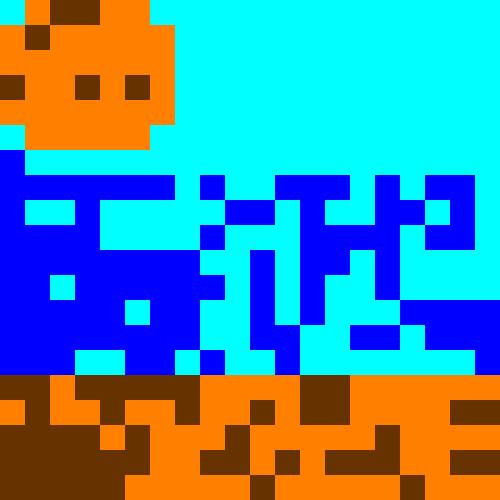}}\,\,
{\includegraphics[width=0.12\linewidth]{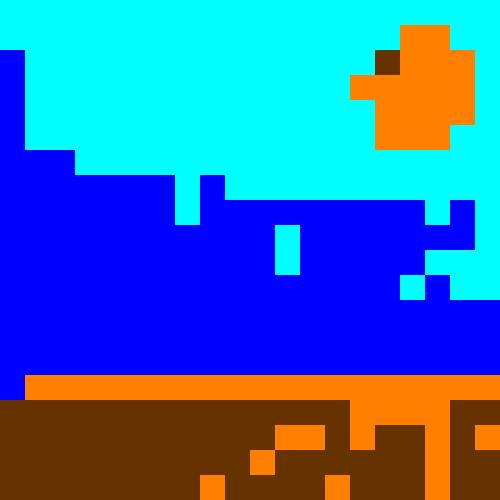}}\,\,
{\includegraphics[width=0.12\linewidth]{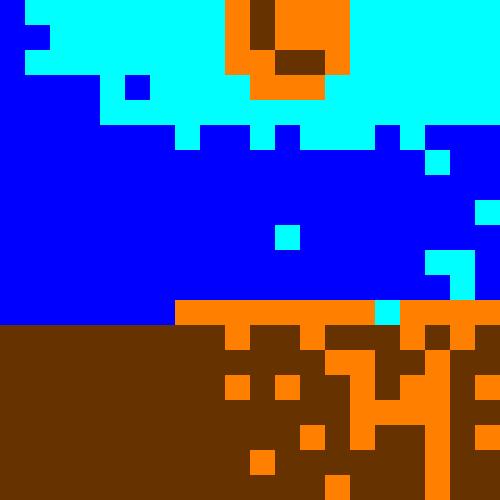}}\,\,
{\includegraphics[width=0.12\linewidth]{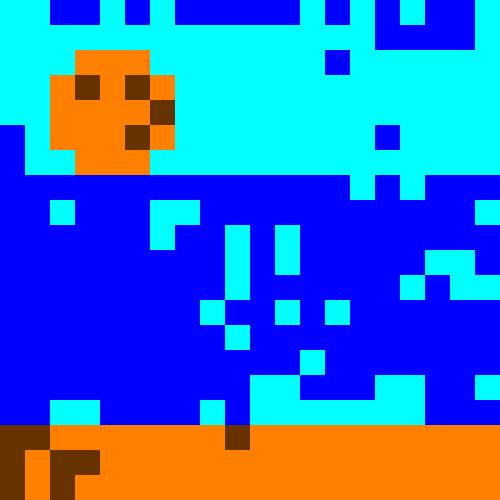}}\,\, \\ \vspace{0.15cm}
{\includegraphics[width=0.12\linewidth]{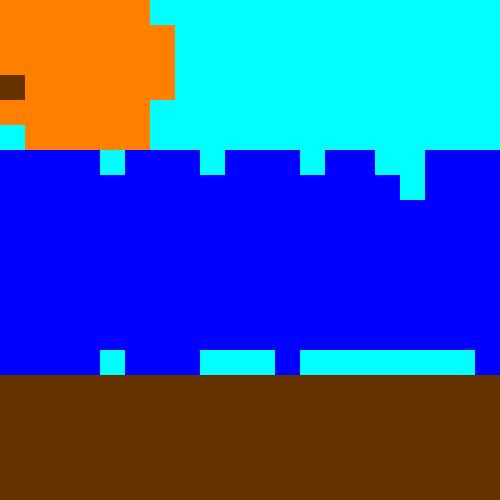}}\,\,
{\includegraphics[width=0.12\linewidth]{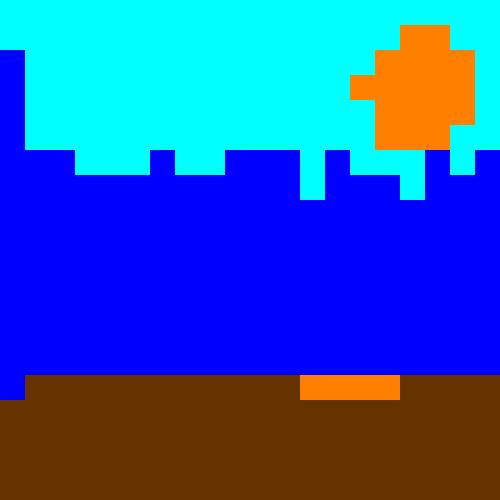}}\,\,
{\includegraphics[width=0.12\linewidth]{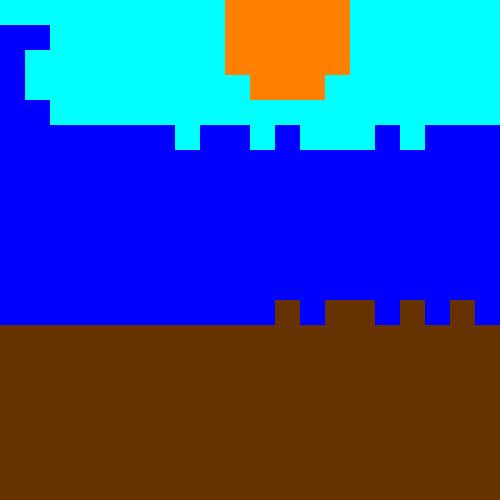}}\,\, 
{\includegraphics[width=0.12\linewidth]{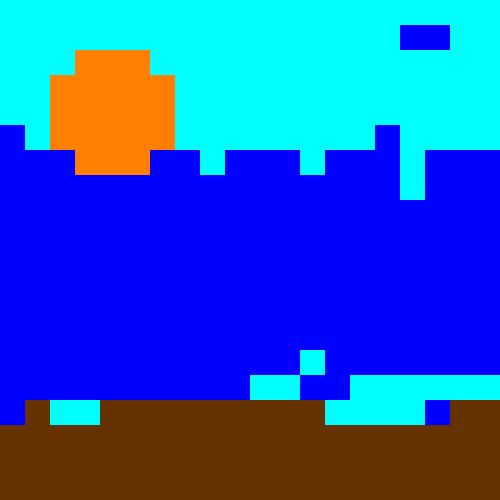}}\,\ \\ \vspace{0.15cm}
{\includegraphics[width=0.12\linewidth]{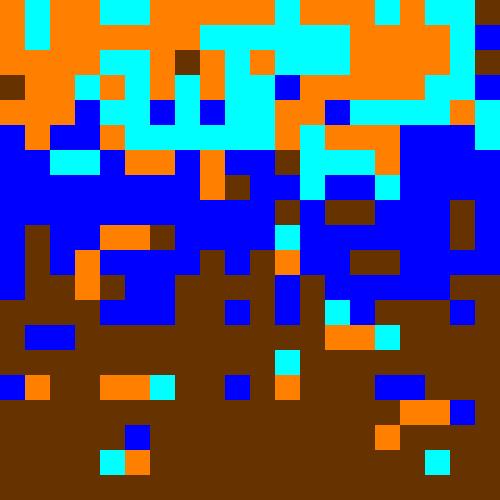}}\,\,
{\includegraphics[width=0.12\linewidth]{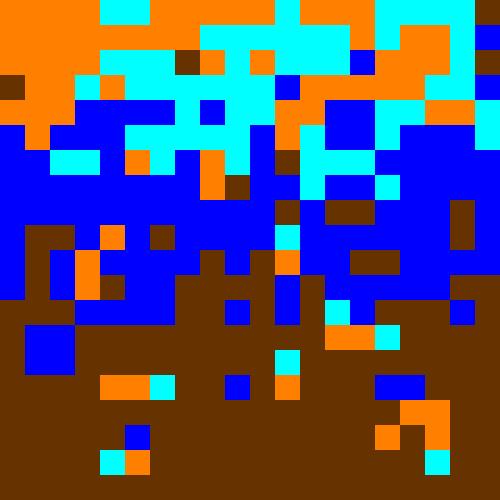}}\,\,
{\includegraphics[width=0.12\linewidth]{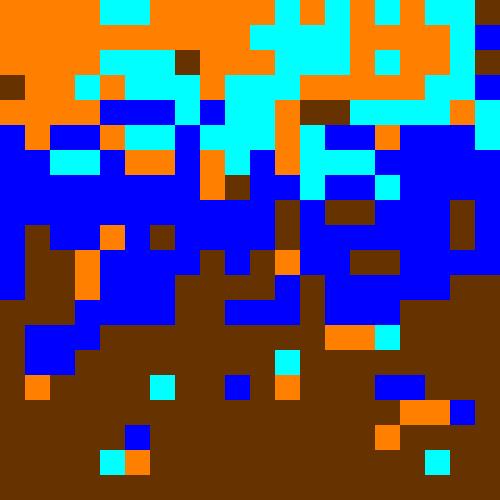}}\,\
{\includegraphics[width=0.12\linewidth]{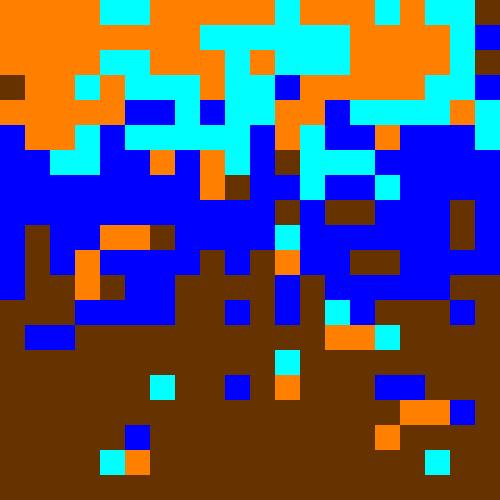}}\,\ \\ \vspace{0.15cm}
{\includegraphics[width=0.12\linewidth]{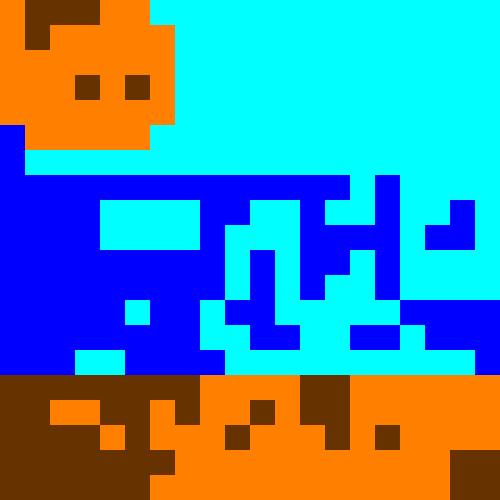}}\,\,
{\includegraphics[width=0.12\linewidth]{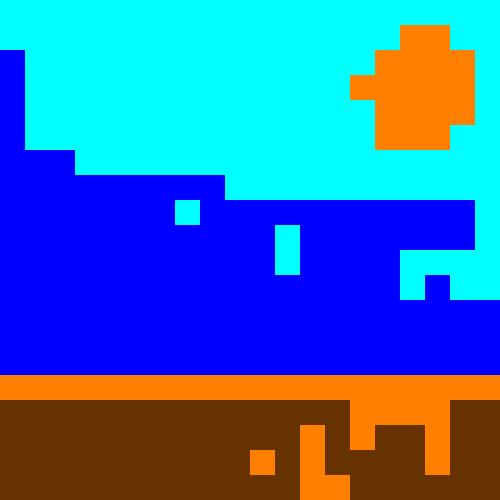}}\,\,
{\includegraphics[width=0.12\linewidth]{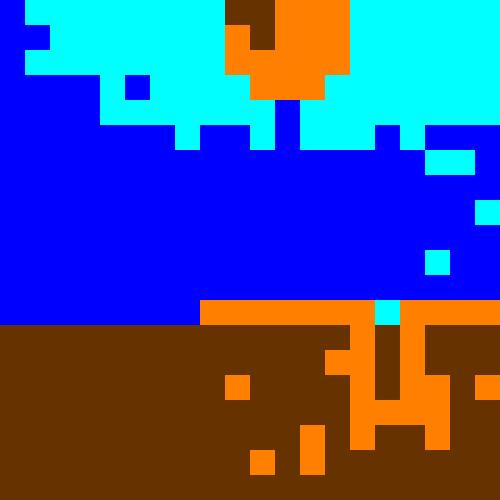}}\,\ 
{\includegraphics[width=0.12\linewidth]{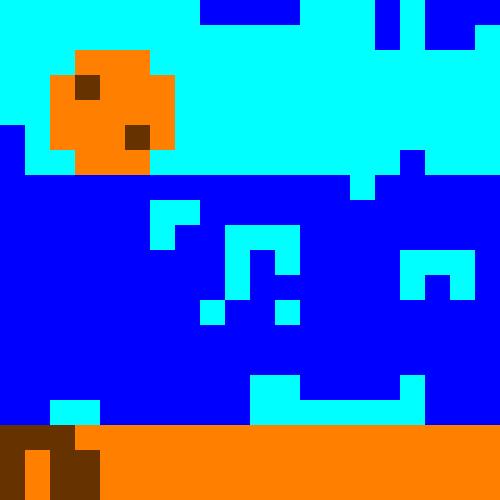}}\,\ \\ \vspace{0.15cm}
{\includegraphics[width=0.12\linewidth]{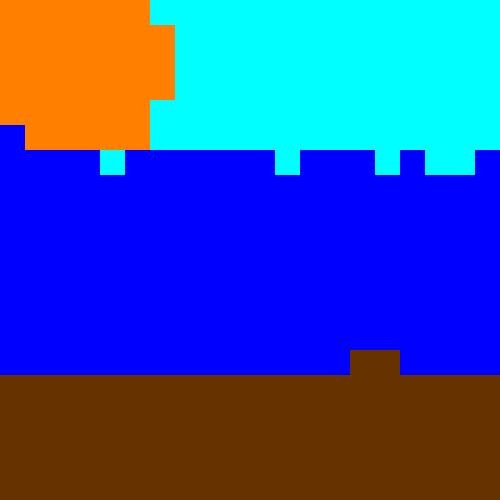}}\,\,
{\includegraphics[width=0.12\linewidth]{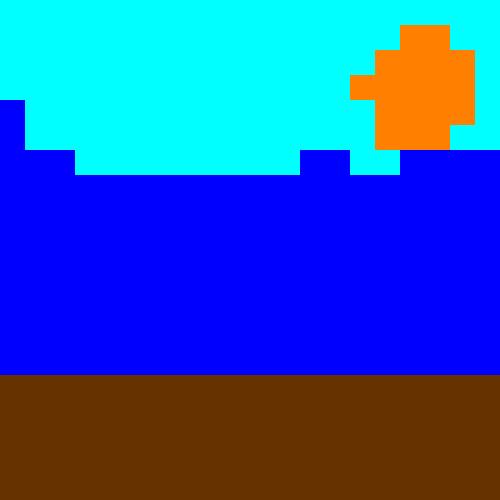}}\,\,
{\includegraphics[width=0.12\linewidth]{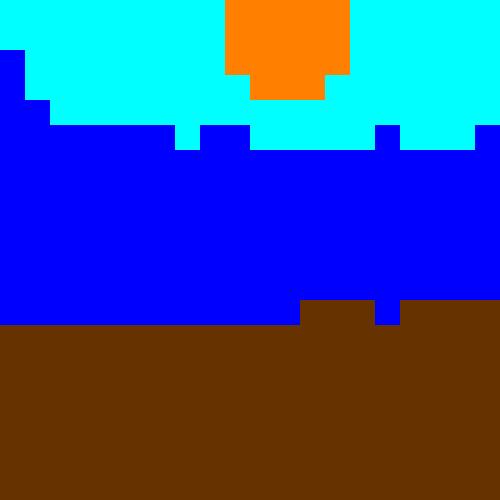}}\,\ 
{\includegraphics[width=0.12\linewidth]{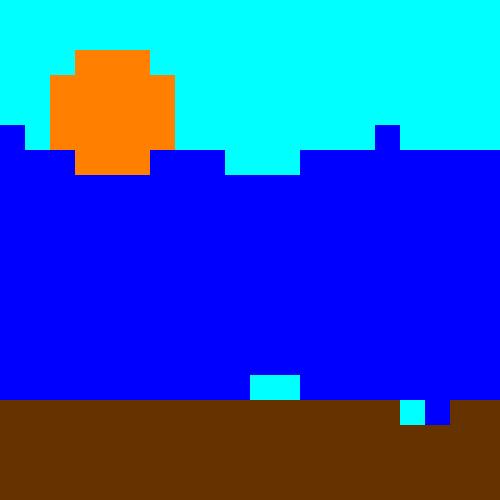}}\,\ \\ \vspace{0.15cm}
{\includegraphics[width=0.12\linewidth]{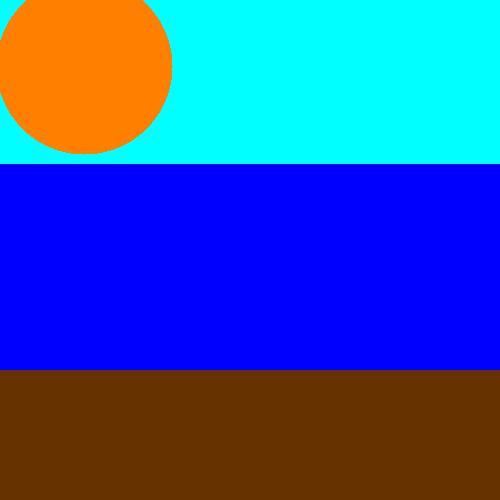}}\,\,
{\includegraphics[width=0.12\linewidth]{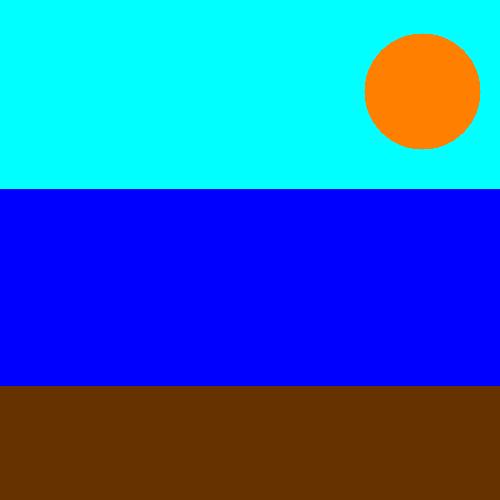}}\,\,
{\includegraphics[width=0.12\linewidth]{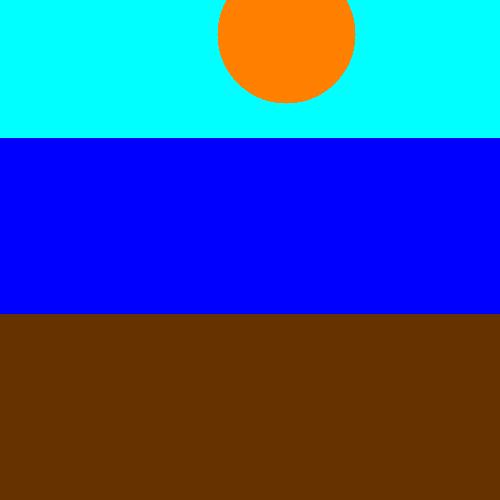}}\,\,
{\includegraphics[width=0.12\linewidth]{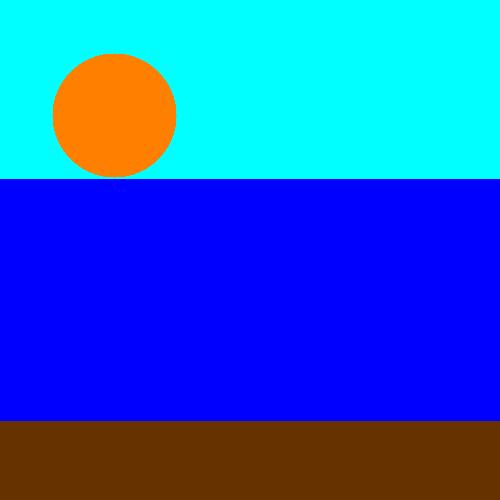}}\,\,\\
\caption{These figures show superpixel segmentation results for different settings including - from 2nd to 7th rows - models 3, 2, 4, 9, 8 and 10 respectively (see line numbers in Table~I). Clean ground truth segmentation blobs are shown in the bottom at the pixel level (``orange'' stands for ``sun'', ``cyan'' for ``sky'', ``blue'' for ``sea'' and ``brown'' for ``sand''.)}\label{images0}
\end{figure}

\noindent {\bf Impact of Dependency and Visual Models.} It is interesting to see that the gain when using the visual model (without dependency model, i.e., lines 2, 5, 8) is more important than the gain obtained when using the dependency model (without visual model, i.e., lines 3, 6, 9), as the former is image/content dependent while the latter acts as a  prior/context. However, it is clear that the gain obtained when combining both models is more substantial and always consistent (lines ``4 vs. 2'', ``7 vs. 5'', ``10 vs. 8''). Note that line 1 is strictly equivalent to void dependency and visual models, and the underlying results correspond to a random classifier, that assigns random class labels to superpixels; the same behavior is observed both with and without the reordering and grouping model. This observation shows that the FSM decoding process is able to benefit from reordering and grouping only if visual and dependency models are able to score the resulting segmentation with a decent precision. The converse is also true as the effect of reordering+grouping on the decoding process is clearly important (lines 5-10 vs. 2-4; see also Fig.~\ref{images0}).

\section{Conclusion} 

In this paper, we introduced a complete framework for scene parsing and annotation based on finite state machines.  The approach  allows us to examine and score many configurations in order to achieve segmentation as well as annotation. Instead of generating intractable configurations of segmentations and labelings, the method relies on finite state machines in order to specify a lattice of these configurations and the best one  (reordering, segmentation, labeling and scoring) corresponds to the shortest path in that lattice. Experiments show that indeed this method is very effective for object class segmentation. 

{
\small
\bibliographystyle{IEEEbib}
\bibliography{refs}

\begin{thebibliography}{10}

\bibitem{Carneiro-Vasconcelos-CVPR05}
G.~Carneiro and N.~Vasconcelos,
\newblock ``Formulating semantic image annotation as a supervised learning
  problem,''
\newblock {\em in Proc. of CVPR}, 2005.

\bibitem{sahbi2003coarse}
H.~Sahbi,
\newblock {\em Coarse-to-fine support vector machines for hierarchical face
  detection},
\newblock Ph.D. thesis, Versailles University, 2003.

\bibitem{tollari2008comparative}
S.~Tollari, P.~Mulhem, M.~Ferecatu, H.~Glotin, M.~Detyniecki, P.~Gallinari,
  H.~Sahbi, and Z-Q. Zhao,
\newblock ``A comparative study of diversity methods for hybrid text and image
  retrieval approaches,''
\newblock in {\em Workshop of the Cross-Language Evaluation Forum for European
  Languages}. Springer, 2008, pp. 585--592.

\bibitem{He2004}
X.~He, R.S. Zimel, and M.A. Carreira,
\newblock ``Multiscale conditional random fields for image labeling,''
\newblock {\em In CVPR}, 2004.

\bibitem{Sin2003}
A.~Singhal, L.~Jiebo, and Z.~Weiyu,
\newblock ``Probabilistic spatial context models for scene content
  understanding,''
\newblock {\em In CVPR}, 2003.

\bibitem{bourdis2011constrained}
N.~Bourdis, D.~Marraud, and H.~Sahbi,
\newblock ``Constrained optical flow for aerial image change detection,''
\newblock in {\em Geoscience and Remote Sensing Symposium (IGARSS), 2011 IEEE
  International}. IEEE, 2011, pp. 4176--4179.

\bibitem{Nowak2010}
S.~Nowak and M.~Huiskes,
\newblock ``New strategies for image annotation: Overview of the photo
  annotation task at imageclef 2010,''
\newblock {\em in The Working Notes of CLEF 2010}, 2010.

\bibitem{boujemaa2004visual}
N.~Boujemaa, F.~Fleuret, V.~Gouet, and H.~Sahbi,
\newblock ``Visual content extraction for automatic semantic annotation of
  video news,''
\newblock in {\em the proceedings of the SPIE Conference, San Jose, CA}, 2004,
  vol.~6.

\bibitem{ref7}
P.~Duygulu, K.~Barnard, J.F.G. deFreitas, and D.~Forsyth,
\newblock ``Object recognition as machine translation: Learning a lexicon for a
  fixed image vocabulary,''
\newblock {\em In: Heyden, A., Sparr, G., Nielsen, M., Johansen, P. (eds.) ECCV
  2002. LNCS, vol. 2353, pp. 97-112. Springer, Heidelberg}, 2002.

\bibitem{napoleon20102d}
T.~Napol{\'e}on and H.~Sahbi,
\newblock ``From 2d silhouettes to 3d object retrieval: contributions and
  benchmarking,''
\newblock {\em Journal on Image and Video Processing}, vol. 2010, pp. 1, 2010.

\bibitem{Jeon-Lavrenko-Manmatha-sigir2003}
J.~Jeon, V.~Lavrenko, and R.Manmatha,
\newblock ``Automatic image annotation and retrieval using cross-media
  relevance models,''
\newblock {\em in Proc. of ACM SIGIR}, pp. 119--126, 2003.

\bibitem{ref18}
J.~Liu, B.~Wang, M.~Li, Z.~Li, W.~Ma, H.~Lu, and S.~Ma,
\newblock ``Dual cross-media relevance model for image annotation,''
\newblock {\em In Proc. of ACM MULTIMEDIA, pp. 605-614}, 2007.

\bibitem{sahbi2002coarse}
H.~Sahbi and N.~Boujemaa,
\newblock ``Coarse-to-fine support vector classifiers for face detection,''
\newblock in {\em Pattern Recognition, 2002. Proceedings. 16th International
  Conference on}. IEEE, 2002, vol.~3, pp. 359--362.

\bibitem{ref14}
V.~Lavrenko, R.~Manmatha, and J.~Jeon,
\newblock ``A model for learning the semantics of pictures,''
\newblock {\em In: Proc. of NIPS}, 2004.

\bibitem{ref8}
S.~Feng, R.~Manmatha, and V.~Lavrenko,
\newblock ``Multiple bernoulli relevance models for image and video
  annotation,''
\newblock {\em In: Proc. of ICCV, pp. 1002-1009}, 2004.

\bibitem{Liu-Li-Liu-pr2009}
J.~Liu, M.~Li, Q.~Liu, H.~Lu, and S.~Ma,
\newblock ``Image annotation via graph learning,''
\newblock {\em Pattern Recognition}, vol. 42, no. 2, pp. 218--228, 2009.

\bibitem{Sahbi1}
H.~Sahbi, J-Y. Audibert, and R.~Keriven,
\newblock ``Context-dependent kernels for object classification,''
\newblock {\em In Pattern Analysis and Machine Intelligence (PAMI)}, vol. 4,
  no. 33, 2011.

\bibitem{Li-Wang-PAMI03}
J.~Li and J.~Z. Wang,
\newblock ``Automatic linguistic indexing of pictures by a statistical modeling
  approach,''
\newblock {\em IEEE Trans. on PAMI}, vol. 25, no. 9, pp. 1075--1088, 2003.

\bibitem{Moser2012}
G.~Moser and B.S. Serpico,
\newblock ``Combining support vector machines and markov random fields in an
  integrated framework for contextual image classification,''
\newblock {\em In TGRS}, 2012.

\bibitem{li2000image}
J.~Li, A.~Najmi, and R-M. Gray,
\newblock ``Image classification by a two-dimensional hidden markov model,''
\newblock {\em IEEE transactions on signal processing}, vol. 48, no. 2, pp.
  517--533, 2000.

\bibitem{choi2001multiscale}
H.~Choi and R-G. Baraniuk,
\newblock ``Multiscale image segmentation using wavelet-domain hidden markov
  models,''
\newblock {\em IEEE Transactions on Image Processing}, vol. 10, no. 9, pp.
  1309--1321, 2001.

\bibitem{Barnard-Duygululu-Forsyth-JMLR03}
K.~Barnard, P.Duygululu, D.~Forsyth, D.~Blei, and M.~Jordan,
\newblock ``Matching words and pictures,''
\newblock {\em JMLR}, 2003.

\bibitem{blei2003latent}
D-M. Blei, A-Y. Ng, and M-I. Jordan,
\newblock ``Latent dirichlet allocation,''
\newblock {\em Journal of machine Learning research}, vol. 3, no. Jan, pp.
  993--1022, 2003.

\bibitem{rasiwasia2013latent}
N.~Rasiwasia and N.~Vasconcelos,
\newblock ``Latent dirichlet allocation models for image classification,''
\newblock {\em IEEE Transactions on Pattern Analysis \& Machine Intelligence},
  , no. 11, pp. 2665--2679, 2013.

\bibitem{wang2008spatial}
X.~Wang and E.~Grimson,
\newblock ``Spatial latent dirichlet allocation,''
\newblock in {\em Advances in neural information processing systems}, 2008, pp.
  1577--1584.

\bibitem{Monay-GaticaPerez-acmmm04}
F.~Monay and D.~GaticaPerez,
\newblock ``Plsa-based image autoannotation: Constraining the latent space,''
\newblock {\em in ACM MM}, 2004.

\bibitem{jin2012image}
B.~Jin, W.~Hu, and H.~Wang,
\newblock ``Image classification based on plsa fusing spatial relationships
  between topics,''
\newblock {\em IEEE Signal Processing Letters}, vol. 19, no. 3, pp. 151--154,
  2012.

\bibitem{zhong2015scene}
Y.~Zhong, Q.~Zhu, and L.~Zhang,
\newblock ``Scene classification based on the multifeature fusion probabilistic
  topic model for high spatial resolution remote sensing imagery,''
\newblock {\em IEEE Transactions on Geoscience and Remote Sensing}, vol. 53,
  no. 11, pp. 6207--6222, 2015.

\bibitem{Sahbi6}
H.~Sahbi and X.~Li,
\newblock ``Context based support vector machines for interconnected image
  annotation (the saburo tsuji best regular paper award),''
\newblock {\em In the Asian Conference on Computer Vision (ACCV)}, 2010.

\bibitem{kim2002support}
K.~Kim, K.~Jung, S-H. Park, and H-J. Kim,
\newblock ``Support vector machines for texture classification,''
\newblock {\em IEEE transactions on pattern analysis and machine intelligence},
  vol. 24, no. 11, pp. 1542--1550, 2002.

\bibitem{Gao-Fan-Xue-Jain-acmmm06}
Y.~Gao, J.~Fan, X.~Xue, and R.~Jain,
\newblock ``Automatic image annotation by incorporating feature hierarchy and
  boosting to scale up svm classifiers,''
\newblock {\em in Proc. of ACM MULTIMEDIA}, 2006.

\bibitem{sahbi2015imageclef}
H.~Sahbi,
\newblock ``Imageclef annotation with explicit context-aware kernel maps,''
\newblock {\em International Journal of Multimedia Information Retrieval}, vol.
  4, no. 2, pp. 113--128, 2015.

\bibitem{wang2011color}
X-Y. Wang, T.~Wang, and B-J.,
\newblock ``Color image segmentation using pixel wise support vector machine
  classification,''
\newblock {\em Pattern Recognition}, vol. 44, no. 4, pp. 777--787, 2011.

\bibitem{vo2012transductive}
P.~Vo and H.~Sahbi,
\newblock ``Transductive kernel map learning and its application to image
  annotation,''
\newblock in {\em BMVC}, 2012, pp. 1--12.

\bibitem{sahbi2013cnrs}
H.~Sahbi,
\newblock ``Cnrs-telecom paristech at imageclef 2013 scalable concept image
  annotation task: Winning annotations with context dependent svms.,''
\newblock in {\em CLEF (Working Notes)}, 2013.

\bibitem{Semo2010}
D.~Semenovich and A.~Sowmya,
\newblock ``Geometry aware local kernels for object recognition,''
\newblock {\em In ACCV}, 2010.

\bibitem{lecun2015deep}
Y.~LeCun, Y.~Bengio, and G.~Hinton,
\newblock ``Deep learning,''
\newblock {\em nature}, vol. 521, no. 7553, pp. 436, 2015.

\bibitem{jiu2017nonlinear}
M.~Jiu and H.~Sahbi,
\newblock ``Nonlinear deep kernel learning for image annotation,''
\newblock {\em IEEE Transactions on Image Processing}, vol. 26, no. 4, pp.
  1820--1832, 2017.

\bibitem{ChenPKMY18}
L-C. Chen, G.~Papandreou, I.~Kokkinos, K.~Murphy, and A-L. Yuille,
\newblock ``Deeplab: Semantic image segmentation with deep convolutional nets,
  atrous convolution, and fully connected crfs,''
\newblock {\em {IEEE} Trans. Pattern Anal. Mach. Intell.}, vol. 40, no. 4, pp.
  834--848, 2018.

\bibitem{krizhevsky2012imagenet}
A.~Krizhevsky, I.~Sutskever, and G.~Hinton,
\newblock ``Imagenet classification with deep convolutional neural networks,''
\newblock in {\em NIPS}, 2012, pp. 1097--1105.

\bibitem{szegedy2015going}
C.~Szegedy, W.~Liu, Y.~Jia, P.~Sermanet, S.~Reed, D.~Anguelov, D.~Erhan,
  V.~Vanhoucke, and A.~Rabinovich,
\newblock ``Going deeper with convolutions,''
\newblock in {\em Proceedings of the IEEE conference on computer vision and
  pattern recognition}, 2015, pp. 1--9.

\bibitem{sahbi2017coarse}
H.~Sahbi,
\newblock ``Coarse-to-fine deep kernel networks,''
\newblock in {\em Computer Vision Workshop (ICCVW), 2017 IEEE International
  Conference on}. IEEE, 2017, pp. 1131--1139.

\bibitem{he2016deep}
K.~He, X.~Zhang, S.~Ren, and J.~Sun,
\newblock ``Deep residual learning for image recognition,''
\newblock in {\em IEEE CVPR}, 2016, pp. 770--778.

\bibitem{resnet2015}
K.~He, X.~Zhang, S.~Ren, and J.~Sun,
\newblock ``Deep residual learning for image recognition,''
\newblock {\em CoRR}, vol. abs/1512.03385, 2015.

\bibitem{badrinarayanan2017segnet}
V.~Badrinarayanan, A.~Kendall, and R.~Cipolla,
\newblock ``Segnet: A deep convolutional encoder-decoder architecture for image
  segmentation,''
\newblock {\em TPAMI}, vol. 39, no. 12, pp. 2481--2495, 2017.

\bibitem{jiu2016deep}
M.~Jiu and H.~Sahbi,
\newblock ``Deep kernel map networks for image annotation,''
\newblock in {\em Acoustics, Speech and Signal Processing (ICASSP), 2016 IEEE
  International Conference on}. IEEE, 2016, pp. 1571--1575.

\bibitem{Ladick-Russell-Kohli-ICCV2009}
L.~Ladicky, C.~Russell, and P.~Kohli,
\newblock ``Associative hierarchical crfs for object class image
  segmentation,''
\newblock {\em in Proc ICCV}, 2009.

\bibitem{Pantofaru-Schmid-Hebert-ECCV2008}
C.~Pantofaru, C.~Schmid, and M.~Hebert,
\newblock ``Object recognition by integrating multiple image segmentations,''
\newblock {\em ECCV}, 2008.

\bibitem{Reynolds-Murphy-CRV2007}
J.~Reynolds and K.~Murphy,
\newblock ``Figure-ground segmentation using a hierarchical conditional random
  field,''
\newblock {\em in Proc. Fourth Canadian Conference on Computer and Robot
  Vision}, 2007.

\bibitem{Sahbi10}
X.~Li and H.~Sahbi,
\newblock ``Superpixel based object class segmentation using conditional random
  fields,''
\newblock {\em In the International Conference on Acoustics, Speech, and Signal
  Processing (ICASSP)}, 2011.

\bibitem{Girshick15}
R-B. Girshick,
\newblock ``Fast {R-CNN},''
\newblock in {\em 2015 {IEEE} International Conference on Computer Vision,
  {ICCV} 2015, Santiago, Chile, December 7-13, 2015}, 2015, pp. 1440--1448.

\bibitem{Shotton-Johnson-Cipolla-CVPR2008}
J.~Shotton, M.~Johnson, and R.~Cipolla,
\newblock ``Semantic texton forests for image categorization and
  segmentation,''
\newblock {\em in Proc. CVPR}, 2008.

\bibitem{Batra-Sukthankar-Tsuhan-CVPR2008}
D.~Batra, R.~Sukthankar, and C.~Tsuhan,
\newblock ``Learning class-specific affinities for image labelling,''
\newblock {\em in Proc. CVPR}, 2008.

\bibitem{Galleguillos-Rabinovich-Belongie-CVPR2008}
C.~Galleguillos, A.~Rabinovich, and S.~Belongie,
\newblock ``Object categorization using co-occurrence, location and
  appearance,''
\newblock {\em in Proc. CVPR}, 2008.

\bibitem{Yang-Meer-Foran-CVPR2007}
L.~Yang, P.~Meer, and D.J. Foran,
\newblock ``Multiple class segmentation using a unified framework over
  mean-shift patches,''
\newblock {\em in Proc CVPR}, 2007.

\bibitem{FelzenszwalbH04}
P-F. Felzenszwalb and D-P. Huttenlocher,
\newblock ``Efficient graph-based image segmentation,''
\newblock {\em International Journal of Computer Vision}, vol. 59, no. 2, pp.
  167--181, 2004.

\bibitem{Shotton-Winn-Rother-A-Criminisi-ECCV2006}
J.~Shotton, J.~Winn, C.~Rother, and A.~Criminisi,
\newblock ``Textonboost: Joint appearance, shape and context modeling for
  multi-class object recognition and segmentation,''
\newblock {\em ECCV}, pp. 1--15, 2006.

\bibitem{Kohli-Ladick-Torr-CVPR2008}
P.~Kohli, L.~Ladicky, and P.H.S. Torr,
\newblock ``Robust higher order potentials for enforcing label consistency,''
\newblock {\em CVPR}, 2008.

\bibitem{sahbi2008robust}
H.~Sahbi, J-Y. Audibert, J.~Rabarisoa, and R.~Keriven,
\newblock ``Robust matching and recognition using context-dependent kernels,''
\newblock in {\em Proceedings of the 25th international conference on Machine
  learning}. ACM, 2008, pp. 856--863.

\bibitem{jiu2016laplacian}
M.~Jiu and H.~Sahbi,
\newblock ``Laplacian deep kernel learning for image annotation,''
\newblock in {\em Acoustics, Speech and Signal Processing (ICASSP), 2016 IEEE
  International Conference on}. IEEE, 2016, pp. 1551--1555.

\bibitem{Wang-Yan-Zhang-Zhang-cvpr2009}
C.~Wang, S.~Yan, L.~Zhang, and H.~Zhang,
\newblock ``Multi-label sparse coding for automatic image annotation,''
\newblock {\em CVPR}, 2009.

\bibitem{fcn2015}
J.~Long, E.~Shelhamer, and T.~Darrell,
\newblock ``Fully convolutional networks for semantic segmentation,''
\newblock {\em CoRR}, vol. abs/1411.4038, 2014.

\bibitem{unets15}
O.~Ronneberger, P.~Fischer, and T.~Brox,
\newblock ``U-net: Convolutional networks for biomedical image segmentation,''
\newblock {\em CoRR}, vol. abs/1505.04597, 2015.

\bibitem{long2015fully}
J.~Long, E.~Shelhamer, and T.~Darrell,
\newblock ``Fully convolutional networks for semantic segmentation,''
\newblock in {\em Proceedings of the IEEE CVPR}, 2015, pp. 3431--3440.

\bibitem{sahbi2014NIPS}
H.~Sahbi,
\newblock ``Parsing images with finite state machines for object class
  segmentation and annotation,''
\newblock {\em Technical Report (N 2013D003), Telecom ParisTech}, 2013.

\bibitem{thiemert2005applying}
S.~Thiemert, H.~Sahbi, and M.~Steinebach,
\newblock ``Applying interest operators in semi-fragile video watermarking,''
\newblock in {\em Security, Steganography, and Watermarking of Multimedia
  Contents VII}. International Society for Optics and Photonics, 2005, vol.
  5681, pp. 353--363.

\bibitem{thiemert2006using}
S.~Thiemert, H.~Sahbi, and M.~Steinebach,
\newblock ``Using entropy for image and video authentication watermarks,''
\newblock in {\em Security, Steganography, and Watermarking of Multimedia
  Contents VIII}. International Society for Optics and Photonics, 2006, vol.
  6072, p. 607218.

\end{thebibliography}
}
 
\end{document}